\documentclass[10pt,twocolumn,letterpaper]{article}

\usepackage{cvpr}              

\usepackage[dvipsnames]{xcolor}

\usepackage{amsmath,amsfonts,amssymb}
\usepackage{algorithmic}
\usepackage{algorithm}
\usepackage{xcolor}

\usepackage{array}
\usepackage{textcomp}
\usepackage{stfloats}
\usepackage{url}
\usepackage{verbatim}

\usepackage{multirow}
\usepackage[normalem]{ulem}
\useunder{\uline}{\ul}{}

\usepackage[accsupp]{axessibility}

\definecolor{cvprblue}{rgb}{0.21,0.49,0.74}
\usepackage[bookmarks=false, pagebackref,breaklinks,colorlinks,citecolor=cvprblue]{hyperref}
\usepackage{microtype}

\def\paperID{12609} 
\def\confName{CVPR}
\def\confYear{2024}

\title{CA-Jaccard: Camera-aware Jaccard Distance for Person Re-identification}

\author{Yiyu Chen, Zheyi Fan\thanks{Corresponding author.}, Zhaoru Chen, Yixuan Zhu\\
Beijing Institute of Technology, China\\
{\tt\small \{yiyuchen1998, Cchenzhaoru\}@gmail.com, \{funye, zhuyixuan\}@bit.edu.cn}
}

\begin{document}
\maketitle
\begin{abstract}
Person re-identification (re-ID) is a challenging task that aims to learn discriminative features for person retrieval. In person re-ID, Jaccard distance is a widely used distance metric, especially in re-ranking and clustering scenarios. However, we discover that camera variation has a significant negative impact on the reliability of Jaccard distance. In particular, Jaccard distance calculates the distance based on the overlap of relevant neighbors. Due to camera variation, intra-camera samples dominate the relevant neighbors, which reduces the reliability of the neighbors by introducing intra-camera negative samples and excluding inter-camera positive samples. To overcome this problem, we propose a novel camera-aware Jaccard (CA-Jaccard) distance that leverages camera information to enhance the reliability of Jaccard distance. Specifically, we design camera-aware k-reciprocal nearest neighbors (CKRNNs) to find k-reciprocal nearest neighbors on the intra-camera and inter-camera ranking lists, which improves the reliability of relevant neighbors and guarantees the contribution of inter-camera samples in the overlap. Moreover, we propose a camera-aware local query expansion (CLQE) to mine reliable samples in relevant neighbors by exploiting camera variation as a strong constraint and assign these samples higher weights in overlap, further improving the reliability. Our CA-Jaccard distance is simple yet effective and can serve as a general distance metric for person re-ID methods with high reliability and low computational cost. Extensive experiments demonstrate the effectiveness of our method. Code is available at \url{https://github.com/chen960/CA-Jaccard/}.
\end{abstract}    
\section{Introduction}
\label{sec:introduction}
\begin{figure}[t]
	\centering
	\includegraphics[width=0.4\textwidth]{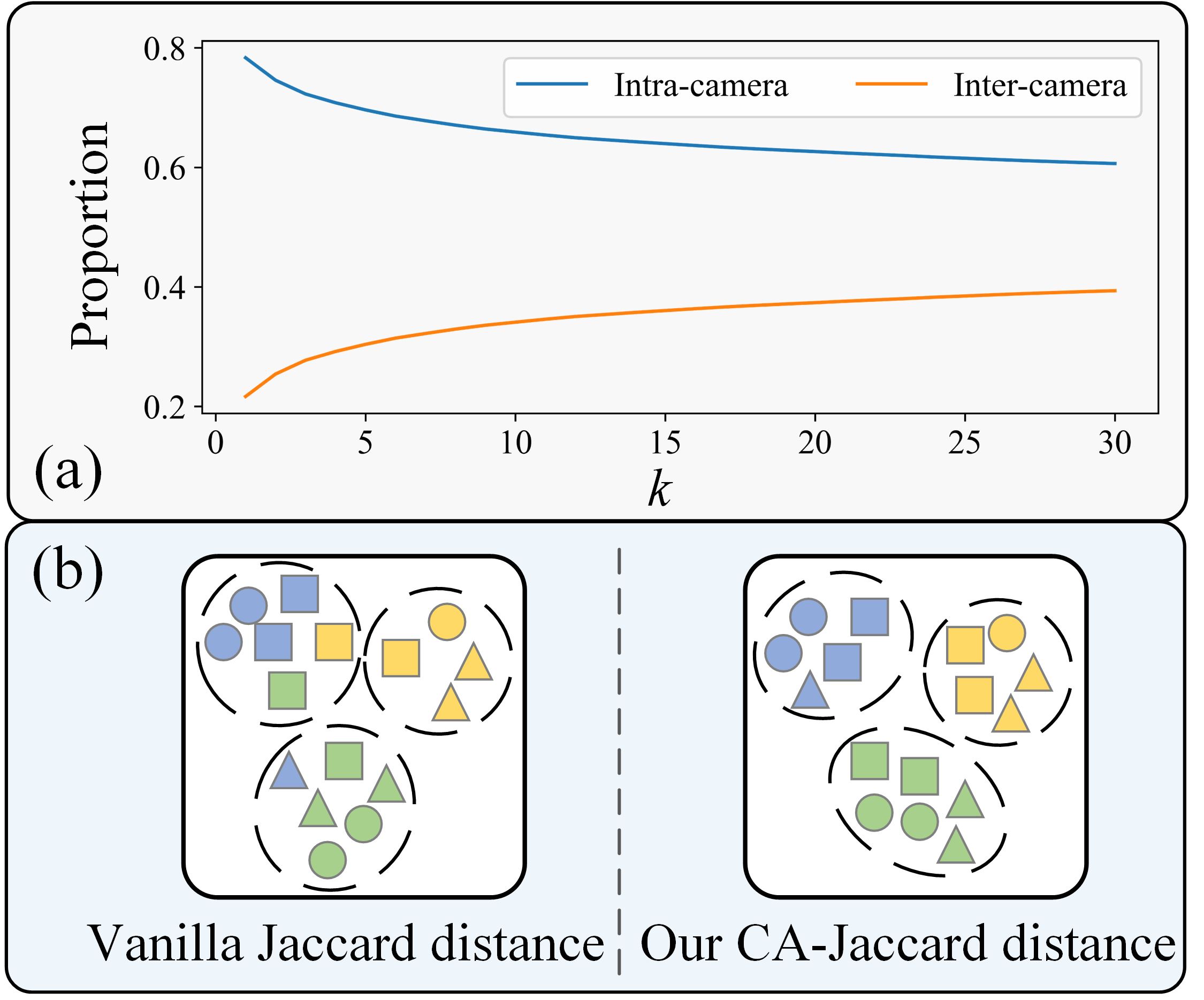}
	\caption{(a) Illustration of the average proportion of intra-camera and inter-camera samples in k-nearest neighbors of all samples. Due to camera variation, the average proportion of intra-camera samples in all samples' k-nearest neighbors is significantly higher than that of inter-camera samples. (b) Comparison of the feature spaces of using Jaccard distance and our CA-Jaccard distance. Different colors represent different identities and different shapes indicate different camera labels.}
	\label{introduction}
\end{figure}
Person re-identification (re-ID) aims to retrieve persons across non-overlapping camera views. It has drawn wide attention due to the growing demand for intelligent surveillance systems. Thanks to the advancement of deep learning, supervised re-ID methods \cite{PCB, BoT, ABD-Net, DG-Net, Nformer} have achieved remarkable performance. However, these methods rely on sufficient person identity annotations, limiting their application in real world scenarios. Hence, recent studies have focused on unsupervised re-ID, seeking to learn discriminative features using unlabeled data.

Recently, most state-of-the-art unsupervised re-ID methods, \ie clustering-based re-ID methods \cite{SpCL, ISE, CC, PAST, SSG}, generally employ a iterative two-stage training procedure: 1) generating pseudo labels based on the Jaccard distance \cite{jaccard} between all training samples using a clustering algorithm \cite{DBSCAN, kmeans, HDBSCAN}; 2) training the re-ID model with the generated pseudo labels. Despite their effectiveness,  these methods still suffer from label noise. To overcome the above problem, numerous approaches \cite{SpCL, RLCC, PPLR, MMT, MEBNet, ACT, Secret} have been proposed. These approaches focus on improving or refining pseudo labels after clustering. 

Moreover, there also have some re-ranking methods \cite{jaccard, SFT, ECN} are proposed to further improve the performance of re-ID methods. K-reciprocal re-ranking \cite{jaccard} is a popular re-ranking method, which utilizes Jaccard distance to re-calculate the distance. 

As mentioned above, Jaccard distance \cite{jaccard} is widely used in person re-ID.
However, Jaccard distance overlooks the detrimental impact of camera variation  (\eg viewpoint, illumination and background), which substantially contributes to label noise in clustering scene and performance degradation in re-ranking scene. Specifically, based on the original distance matrix (\ie Euclidean distance or cosine distance), Jaccard distance measures the distance between samples based on the overlap of their relevant neighbors, which means the accuracy of relevant neighbors determines the reliability of Jaccard distance. The relevant neighbors are found by applying k-reciprocal nearest constraint and encoded into a weighted unit vector called weighted neighbors vector. Higher weights are assigned to closer neighbors to reflect their greater contribution to the overlap calculation. However, as shown in Fig. \ref{introduction} (a), due to camera variation, intra-camera samples dominate the k-nearest neighbors, resulting high proportion and weight of intra-camera samples in the weighted relevant neighbors vectors. It undermines the reliability of Jaccard distance by introducing many high weight intra-camera negative samples and hindering informative inter-camera positive samples into weighted relevant neighbors vectors. Moreover, Jaccard distance utilizes local query expansion to expand weighted relevant neighbors vector of a sample by averaging the weighted relevant neighbors vectors of its k-nearest neighbors. Since k-nearest neighbors mainly consist of intra-camera samples, the proportion and weight of intra-camera samples are further increased, while the reliability of Jaccard distance is further decreased.

To address these problems, we propose camera-aware Jaccard (CA-Jaccard) distance, a simple yet effective distance metric that enhances the reliability of the Jaccard distance \cite{jaccard} with camera information for more accurate pseudo label generation, which is shown in Fig. \ref{introduction} (b). In particular, our approach modifies the robust k-reciprocal nearest neighbors (KRNNs) and local query expansion (LQE) of Jaccard distance in a camera-aware manner to increase the accuracy of relevant neighbors. We discover that inter-camera samples have more information and reliability. Therefore, to include more inter-camera samples into relevant neighbors and restrain the proportion and weight of intra-camera samples under camera variation, we propose camera-aware k-reciprocal nearest neighbors (CKRNNs) for more accurate relevant neighbors. CKRNNs impose the k-reciprocal nearest constraint separately for the intra-camera and inter-camera ranking lists with different $k$ values, and then combine the neighbors obtained from both. Additionally, to further improve the accuracy of relevant neighbors, we propose camera-aware local query expansion (CLQE) to obtain weighted expanded neighbors vectors by averaging the weighted CKRNNs vectors of intra-camera and inter-camera k-nearest neighbors.  CLQE exploits camera variation as a strong constraint to mine reliable samples that frequently appear in the relevant neighbors of both intra-camera and inter-camera k-nearest neighbors, and enlarges their weights for greater contribution in overlap.

Our contributions can be summarized as follows:

(1) We propose a novel camera-aware Jaccard (CA-Jaccard) distance that leverages camera-aware k-reciprocal nearest neighbors (CKRNNs) and camera-aware local query expansion (CLQE) to enhance the reliability of Jaccard distance.

(2) Our CA-Jaccard distance is simple yet effective, with higher reliability and lower computational cost than Jaccard distance, and can serve as a general distance metric for person re-ID.

(3) Extensive experiments on different person re-ID scenarios demonstrate the effectiveness of our CA-Jaccard distance.
\section{Related work}
\label{sec:relatedwork}

\subsection{Clustering for Unsupervised Person Re-ID}
In unsupervised person re-ID, datasets lack identity label information. Many works utilize clustering \cite{SSG, PAST,PUL, BUC, HCT, ADcluster, ADTC} and k-nearest neighbors \cite{MMCL,ECN_unsupervised, SSL} to generate pseudo labels. 
Clustering-based methods \cite{CC,SSG, PAST} demonstrate their superiority by achieving state-of-the-art performance. They generally leverage the Jaccard distance \cite{jaccard} to compute the distance matrix and then adopt the DBSCAN clustering \cite{DBSCAN} algorithm for pseudo label generation. However, the generated pseudo labels inevitably contain label noise, which severely affects the performance. Recent methods tackle this problem using robust clustering techniques \cite{SpCL, ADcluster}, label refinement procedures\cite{RLCC, PPLR, Secret}, co-teaching algorithms \cite{MMT, MEBNet, ACT, NRMT}. Although these methods strive to reduce label noise, they neglect the label noise caused by unreliable Jaccard distance. RIDES \cite{ride} improves the original distance by reducing the distance of reliable inter-camera sample pairs, which improves the accuracy of relevant neighbors and the reliability of Jaccard distance implicitly and limitedly.
Different from \cite{ride}, our method enhances Jaccard distance directly, and improves the accuracy of relevant neighbors significantly and stably.
In this paper, our method brings more reliable pseudo labels in clustering scene. 

\subsection{Re-ranking for Person Re-ID}
Re-ranking is a post-processing technique to improve the original retrieval results by the information of near neighbors. In \cite{kNN}, k-nearest neighbors are first used for re-ranking. Many works \cite{jaccard, DaF, ECN} further discover more potential information based on k-nearest neighbors. To reduce the false positives in the top-k of original ranking lists, k-reciprocal nearest neighbors \cite{CDM, HN} are introduced into person re-ID by K-reciprocal(KR) re-ranking\cite{jaccard}. Inspired by sparse contextual activation (SCA) encoding \cite{SCA} and average query expansion (AQE) \cite{AQE}, KR re-ranking \cite{jaccard} searches k-reciprocal neighbors and computes the Jaccard distance with k-reciprocal encoding and local query expansion. ECN \cite{ECN} improves the original pairwise distance by aggregating the distances between expanded neighbors of image pairs.
These methods can not handle the large camera variation in ranking lists well, which significantly hinders the performance of re-ranking. To solve this problem, we attempt to aggregate the camera information into re-ranking.
\section{Methodology}

\begin{figure*}
	\centering
	\includegraphics[width=\textwidth]{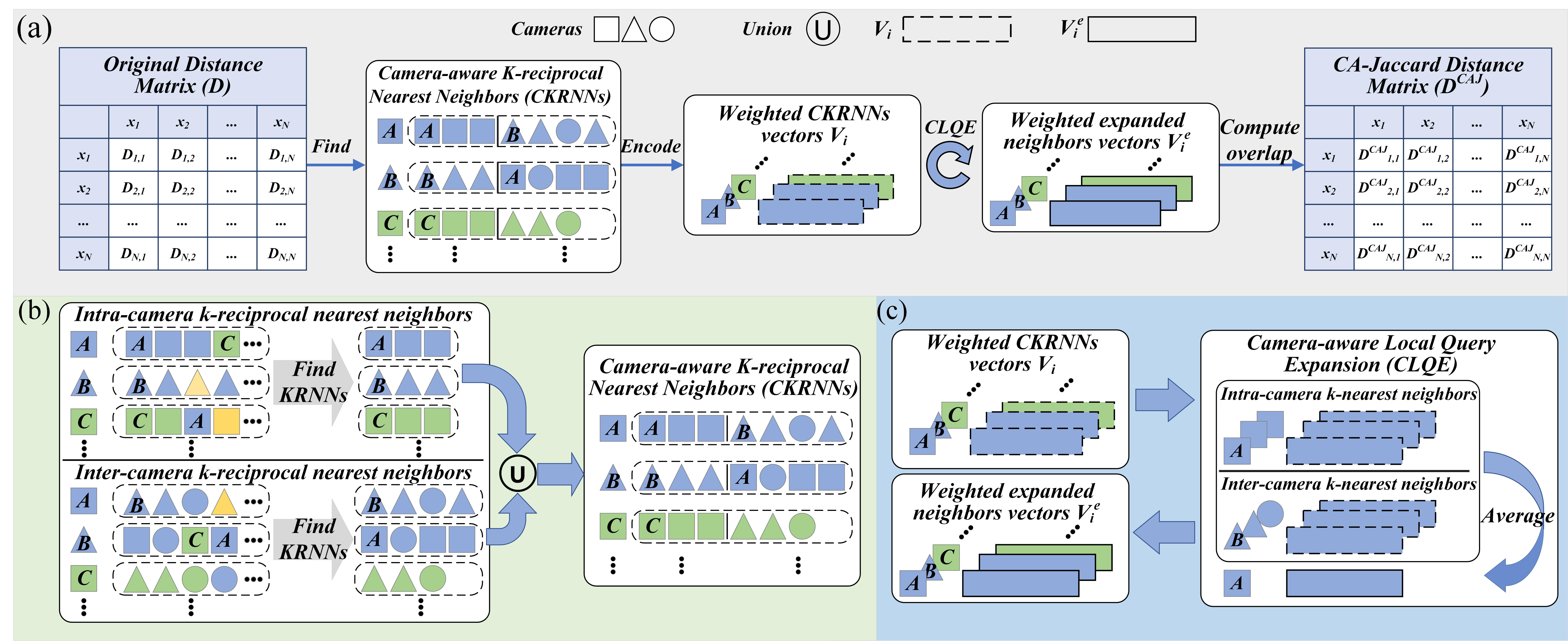}
	\caption{
	Schematic diagram of CA-Jaccard distance. 
	(a) Overview of CA-Jaccard distance computation. Given the original distance matrix, find CKRNNs and encode them as vectors, then use CLQE to generate weighted expanded neighbors vectors. Finally, calculate the overlap between these vectors to obtain the CA-Jaccard distance matrix. 
	(b) CKRNNs. CKRNNs find reliable relevant neighbors by applying the k-reciprocal nearest constraint on intra-camera and inter-camera ranking lists.
	(c) CLQE. CLQE averages the weighted CKRNNs vectors of intra-camera and inter-camera k-nearest neighbors to obtain weighted expanded neighbors.}
	\label{overview}
\end{figure*}

In this section, we first revisit Jaccard distance. Then, we elaborate on the details of our camera-aware Jaccard (CA-Jaccard) distance, which enhances the Jaccard distance by using camera-aware k-reciprocal nearest neighbors (CKRNNs) and camera-aware local query expansion (CLQE). 

\subsection{Preliminary}

Our goal is to compute general and reliable distances between samples for different person re-ID scenarios. The computation procedures of CA-Jaccard distance in the re-ranking and clustering scene are very similar. Therefore, for similarity, we introduce our CA-Jaccard distance within the clustering scenario of clustering-based unsupervised re-ID methods. In this case, we are provided with an unlabeled re-ID training dataset $X = \left\{ {{x_i}} \right\}_{i = 1}^N$ with $N$ images, where $x_i$ denotes the $i$-th image. Each image $x_i$ is associated with its camera label $c_i$.

\subsection{Revisit Jaccard Distance}
The core idea of Jaccard distance is that if two images are similar, their relevant neighbors should also be similar. Based on this assumption, Jaccard distance measures the distance between samples according to the overlap of their relevant neighbors. Jaccard distance incorporates the robust k-reciprocal nearest neighbors (KRNNs) into relevant neighbors and then expands them using local query expansion (LQE). Since neighbors sets treat each neighbor equally and overlap computation of sets is time-consuming, 
Jaccard distance encodes the neighbors sets of samples into weighted unit vectors and transforms set comparison problem into pure vector computation. The detailed calculation steps of Jaccard distance are as follows.

\textbf{Original distance computation.}
The original distance matrix $D$ is obtained by applying either the cosine or Euclidean distance based on the features extracted by the model $f_{\theta}(\cdot)$ from all samples.

\textbf{Robust k-reciprocal nearest neighbors.}
For sample $x_i$, the ranking list ${L_i} = \{ x_1^i,x_2^i,...,x_N^i\}$ can be obtained by arranging samples according to the original distance between $x_i$ and all training samples. The k-nearest neighbors $N({x_i},k)$ of $x_i$ are defined as the top-k samples of ranking list $L_i$:

\begin{equation}
	N({x_i},k) = L_{i}[1:k].
\end{equation}
Then the KRNNs $R({x_i},{k_1})$ can be found:
\setlength{\abovedisplayskip}{3pt}
\setlength{\belowdisplayskip}{3pt}
\begin{equation}
	R({x_i},{k_1}) = \{ {x_j}|{x_i} \in N({x_j},{k_1}) \wedge {x_j} \in N({x_i},{k_1})\}.
\end{equation}
To recall some positive samples may be excluded from the KRNNs, robust KRNNs are computed as follows:
\begin{equation}
	\begin{array}{l}
		{R^*}({x_i},{k_1}) \leftarrow R({x_i},{k_1}) \cup R\left( {{x_j},\frac{1}{2}k_1} \right)\\
		{s.t.}\left| {R({x_i},{k_1}) \cap R\left( {{x_j},\frac{1}{2}{k_1}} \right)} \right|\ge\frac{2}{3}\left| {R\left( {{x_j},\frac{1}{2}{k_1}} \right)} \right|\\
		\forall \,{x_j} \in R({x_i},k_1),
	\end{array}
\end{equation}
where $|\cdot|$ denotes the number of samples in the set. This operation employs a strict constraint to ensure that most of the recalled samples are positive samples. 

\textbf{Vectorization of neighbors.}
To reduce the computational complexity and increase the discriminability of neighbors, the robust KRNNs of sample $x_i$ are encoded into a weighted robust KRNNs vector $V_i = [{V_{i,1}, V_{i,2}, ...,V_{i,N}}]$, where $V_i$ is a $N$-dimension unit vector and $V_{i,j}$ is computed according to the original distance between $x_i$ and $x_j$ if $x_j$ is within the robust k-reciprocal nearest neighbor of $x_i$, otherwise it is zero:
\begin{equation}
	\label{eqn4}
	{V_{i,j}} = \left\{ {\begin{array}{*{20}{l}}
			{\frac{{{{e}^{ - {D_{i,j}}}}}}{{\sum\limits_{{x_l} \in {R^*}\left( {{x_i},{k_1}} \right)} {{{e}^{ - {D_{i,l}}}}} }}}&{if\,x_j \in {R^*}\left( {{x_i},{k_1}} \right)}\\
			0&{{otherwise}},
	\end{array}} \right.
\end{equation}
where $D_{i,j}$ is the original distance between $x_i$ and $x_j$. 

\textbf{Local query expansion.}
Considering similar samples may share similar features and neighbors, LQE is adopted to generate the weighted expanded neighbors vector $V^e_i$ by averaging the weighted robust KRNNs vectors of $x_i$'s k-nearest neighbors:
\begin{equation}
	{V^e_i} = \frac{1}{{|N({x_i},{k_2})|}}\sum\limits_{{x_j} \in N({x_i},{k_2})} {{V_j}}, 
\end{equation}
where $k_{2}<k_{1}$ because there are noise in k-nearest neighbors, $V_j$ denotes the weighted robust KRNNs vector of $x_j$.  

\textbf{Overlap computation.}
The Jaccard distance $D_{i,j}^J$ between $x_i$ and $x_j$ are computed by vectorized overlap computation:
\begin{equation}
	\label{eqn8}
	D_{i,j}^J = 1 - \frac{{\sum\limits_{l = 1}^N {\min } \left( {{V^e_{i,l}},{V^e_{j,l}}} \right)}}{{\sum\limits_{l = 1}^N {\max } \left( {{V^e_{i,l}},{V^e_{j,l}}} \right)}},
\end{equation}
where $min$ and $max$ can be regarded as the intersection and union operation in vector form.

The Jaccard distance is widely used in many methods, but it still has drawbacks. Camera variation makes it difficult for robust KRNNs and LQE to obtain reliable relevant neighbors for overlap computation, which hinders the reliability of Jaccard distance. Therefore, the key motivation of our method is to improve the reliability of relevant neighbors. To achieve this goal, we propose CKRNNs and CLQE to make efforts from different aspects.

\subsection{Camera-aware K-reciprocal Nearest Neighbors}
Although robust KRNNs utilize some constraints to find relevant neighbors, the neighbors are still unreliable. Camera variation causes intra-camera samples to have a high proportion and low ranks in k-nearest neighbors. Consequently, they have a high proportion in robust KRNNs. Negative samples from the same camera are heavily included in robust KRNNs vectors with large weight, while informative and reliable inter-camera samples are hardly included, reducing the reliability of the neighbors. To find more inter-camera relevant samples and restrain the proportion and weight of intra-camera samples, we propose camera-aware k-reciprocal nearest neighbors (CKRNNs), as shown in Fig. \ref{overview}(b).

For sample $x_i$, we obtain the intra-camera ranking list $L_i^{intra}$ and inter-camera ranking list $L_i^{inter}$:
\begin{equation}
	L_i^{intra} = \{ x_1^{i\_intra},x_2^{i\_intra},...x_{ {N_{{c_i}}}}^{i\_intra} \}, 
\end{equation}
\begin{equation}
	L_i^{inter} = \{ x_1^{i\_inter},x_2^{i\_inter},...x_{N - {N_{{c_i}}}}^{i\_inter} \},
\end{equation}
where $x_j^{i\_intra}$  and $x_j^{i\_inter}$ represent the $j$-th sample in the intra-camera and inter-camera ranking list, and $N_{{c_i}}$ means the number of samples share the same camera label $c_i$ as $x_i$. 

Then we find k-nearest neighbors in both ranking lists to obtain intra-camera k-nearest neighbors $N^{intra}({x_i},k_1^{intra})$ and inter-camera k-nearest neighbors $N^{inter}({x_i},k_1^{inter})$:
\begin{equation}
	N^{intra}({x_i},k_1^{intra})  =  {L_i^{intra}}[1:k_1^{intra}],
\end{equation}
\begin{equation}
	N^{inter}({x_i},k_1^{inter})  =  {L_i^{inter}}[1:k_1^{inter}],
\end{equation}
where $k_1^{intra}$ and $k_1^{inter}$ mean different $k$ are used in intra-camera and inter-camera ranking lists.

Next, we impose the k-reciprocal nearest constraint on both intra-camera and inter-camera k-nearest neighbors, and union the obtained neighbors as CKRNNs ${R^c}({x_i},k_1^{intra},k_1^{inter})$, which can be formulated as:
\begin{equation}
	\label{eqn11}
	\begin{aligned}
		&{R^c}({x_i},k_1^{intra},k_1^{inter}) =\\ &\{ {x_j}|{x_i} \in N^{intra}({x_j},k_1^{intra}) \wedge {x_j} \in N^{intra}({x_i},k_1^{intra})\}  \\& \cup \{ {x_j}|{x_i} \in N^{inter}({x_j},k_1^{inter}) \wedge {x_j} \in N^{inter}({x_i},k_1^{inter})\}.
	\end{aligned}
\end{equation}
By using a smaller $k_1^{intra}$, we can include only intra-camera positive samples and exclude intra-camera negative samples. We discover that inter-camera samples are more informative and reliable in overlap computation. Thus, we use a large $k_1^{inter}$ to find more inter-camera samples, increasing the proportion of inter-camera samples in CKRNNs.  When CKRNNs are encoded into a weighted CKRNNs vector, although the weight of each intra-camera sample is relatively large due to their small original distance, a large amount of inter-camera samples in CKRNNs ensure the proportion and total weight of inter-camera samples, which enhances the reliability of neighbors.

Note that we do not use recall operation which has a great positive effect on the reliability of robust KRNNs and Jaccard distance. This is because the key of the recall operation is to recall more inter-camera positive samples, which is explicitly achieved in CKRNNs by applying the k-reciprocal nearest constraint on the inter-camera ranking list.

\subsection{Camera-aware Local Query Expansion}
LQE is used in Jaccard distance to incorporate more samples from the robust KRNNs of k-nearest neighbors and reassign weights of neighbors by averaging weighted robust KRNNs vectors of k-nearest neighbors. Due to camera variation, most k-nearest neighbors are intra-camera samples, which also have a high proportion of intra-camera samples with high weights in their weighted robust KRNNs vectors. As a result, LQE reassigns higher weights to intra-camera negative samples which frequently occur in robust KRNNs of k-nearest neighbors, while reassigning lower weights to inter-camera positive samples which have low proportion but are informative and reliable. In this case, the unreliability of relevant neighbors is further exacerbated.

Unlike LQE which reduces the reliability of neighbors, we propose camera-aware local query expansion (CLQE) to boost the reliability of neighbors in a clever way. As shown in Fig. \ref{overview}(c), CLQE averages the weighted CKRNNs vectors of intra-camera and inter-camera k-nearest neighbors to obtain weighted expanded neighbors:
\begin{equation}
	\label{eqn12}
	\begin{aligned}
		{V^e_i} =& \frac{1}{{|{N^{intra}}({x_i},{k_2^{intra}})| + |{N^{inter}}({x_i},{k_2^{inter}})|}}\times \\
		&(\sum\limits_{{x_j} \in {N^{intra}}({x_i},{k_2^{intra}})} {{V_j}}  + \sum\limits_{{x_l} \in {N^{inter}}({x_i},{k_2^{inter}})} {{V_l}} ),
	\end{aligned}
\end{equation}
where $k_2^{intra}$ and $k_2^{inter}$ are the $k$ number of k-nearest neighbors we select from intra-camera and inter-camera ranking lists, $V_j$ and $V_l$ are the weighted CKRNNs vectors of $x_j$ and $x_l$ respectively. CLQE regards camera variation as a strong constraint to mine reliable samples in neighbors and enlarge their weights. Specifically, CLQE averages the weighted CKRNNs vectors of samples from multiple cameras. Due to the existence of camera variation, the reliability of a sample increases with its frequency of occurrence in the CKRNNs of samples from multiple cameras, indicating that it is more likely to be a positive sample. In this way, CLQE assigns reliable samples that have high frequency in the CKRNNs of intra-camera and inter-camera k-nearest neighbors higher weight.

\subsection{Camera-aware Jaccard Distance}
We name the proposed distance metric camera-aware Jaccard (CA-Jaccard) distance, which improves the reliability of Jaccard distance by replacing the robust KRNNs and LQE with CKRNNs and CLQE. We utilize CKRNNs to increase the proportion and total weight of inter-camera samples which are informative and exclude the intra-camera samples beyond relevant neighbors, enhancing the reliability of neighbors. Meanwhile, we utilize CLQE to assign high weights to reliable samples, which further improves the reliability of relevant neighbors. Our CA-Jaccard distance is a simple but effective distance metric, offering lower computational complexity and higher reliability than Jaccard distance. 
The detailed steps of CA-Jaccard distance computation are presented in Fig. \ref{overview}(a).

\begin{table*}[ht]
	\caption{Comparison with the state-of-the-art unsupervised re-ID methods on Market1501, MSMT17 and VeRi-776. The best results are in \textbf{bold} and the second-best results are in {\ul underline}. CC* denotes our results with the official CC code without hard instance memory updating mechanism and generalized mean (GeM) pooling \cite{GeM}. ``CAJ” represents CA-Jaccard distance.}
	\renewcommand{\arraystretch}{0.9}
	\renewcommand\tabcolsep{5pt}
	\centering
	\begin{tabular}{c|c|cccc|cccc|cccc}
		\hline
		\multirow{2}{*}{Methods} & \multirow{2}{*}{Reference} & \multicolumn{4}{c|}{Market1501}                                & \multicolumn{4}{c|}{MSMT17}                                    & \multicolumn{4}{c}{VeRi-776}                                  \\ \cline{3-14} 
		&                            & mAP           & R1            & R5            & R10           & mAP           & R1            & R5            & R10           & mAP           & R1            & R5            & R10           \\ \hline
		MMCL~\cite{MMCL}         & CVPR'20                    & 45.5          & 80.3          & 89.4          & 92.3          & 11.2          & 35.4          & 44.8          & 49.8          & -             & -             & -             & -             \\
		HCT~\cite{HCT}           & CVPR'20                    & 56.4          & 80.0          & 91.6          & 95.2          & -             & -             & -             & -             & -             & -             & -             & -             \\
		GCL~\cite{GCL}           & CVPR'21                    & 66.8          & 87.3          & 93.5          & 95.5          & 21.3          & 45.7          & 58.6          & 64.5          & -             & -             & -             & -             \\
		IICS~\cite{IICS}         & CVPR'21                    & 72.9          & 89.5          & 95.2          & 97.0          & 26.9          & 56.4          & 68.8          & 73.4          & -             & -             & -             & -             \\
		SpCL~\cite{SpCL}         & NeurIPS'20                 & 73.1          & 88.1          & 95.1          & 97.0            & 19.1          & 42.3          & 55.6          & 61.2          & 36.9          & 79.9          & 86.8          & 89.9          \\
		RLCC~\cite{RLCC}         & CVPR'21                    & 77.7          & 90.8          & 96.3          & 97.5          & 27.9          & 56.5          & 68.4          & 73.1          & 39.6          & 83.4          & 88.8          & 90.9          \\
		OPLG-HCD~\cite{OPLG-HCD} & ICCV'21                    & 78.1          & 91.1          & 96.4          & 97.7          & 26.9          & 53.7          & 65.3          & 70.2          & -             & -             & -             & -             \\
		MCRN~\cite{MCRN}         & AAAI'22                    & 80.8          & 92.5          & –             & -             & 31.2          & 63.6          & -             & -             & -             & -             & -             & -             \\
		Secret~\cite{Secret}     & AAAI'22                    & 81.0          & 92.6          & -             & -             & 31.3          & 60.4          & -             & -             & -             & -             & -             & -             \\
		CC~\cite{CC}             & ACCV'22                    & 82.6          & 93.0          & 97.0          & 98.1          & 33.3          & 63.3          & 73.7          & 77.8          & 42.5          & 87.7          & 91.4          & 93.1          \\
		RESL~\cite{RESL}         & AAAI'22                    & 83.1          & 93.2          & 96.8          & 98.0          & 33.6          & 64.8          & 74.6          & 79.6          & -             & -             & -             & -             \\
		RIDE~\cite{ride}         & SCIS'23                    & 84.0          & 93.0          & 97.3          & -             & 39.5          & 68.4          & 79.6          & -             & -             & -             & -             & -             \\
		ISE~\cite{ISE}           & CVPR'22                    & 84.7          & 94.0          & 97.8          & 98.8          & 35.0          & 64.7          & 75.5          & 79.4          & -             & -             & -             & -             \\ \hline
		CAP~\cite{CAP}           & AAAI'21                    & 79.2          & 91.4          & 96.3          & 97.7          & 36.9          & 67.4          & 78.0          & 81.4          & 40.4          & 86.8          & 90.8          & 92.7          \\ 
		CAP~\cite{CAP}+CAJ       & -                          & 80.4          & 91.7          & 96.4          & 97.7          & 39.9          & 70.0          & 80.5          & 83.7          & 43.4          & 90.4          & 93.4          & 95.1          \\ \hline
		CC*~\cite{CC}            & ACCV'22                    & 81.0          & 91.1          & 96.2          & 97.4          & 31.1          & 60.2          & 71.3          & 75.7          & 38.1          & 80.3          & 85.1          & 87.5          \\
		CC*~\cite{CC}+CAJ        & -                          & {\ul 84.8}    & 93.6          & 97.6          & 98.4          & 42.8          & 72.3          & 82.2          & 85.6          & 43.1          & 90.1          & 92.8          & 95.0          \\ \hline
		ICE~\cite{ICE}           & ICCV'21                    & 82.3          & 93.8          & 97.6          & 98.4          & 38.9          & 70.2          & 80.5          & 84.4          & 42.5          & 87.5          & 91.5          & 93.2          \\
		ICE~\cite{ICE}+CAJ       & -                          & 82.7          & 93.8          & 97.7          & 98.4          & {\ul 43.0}    & {\ul 74.1}    & {\ul 83.8}    & {\ul 86.9}    & {\ul 44.5}    & \textbf{91.0} & {\ul 93.6}    & {\ul 95.0}    \\ \hline
		PPLR~\cite{PPLR}         & CVPR'22                    & 84.4          & {\ul 94.3}    & {\ul 97.8}    & {\ul 98.6}    & 42.2          & 73.3    & 83.5    & 86.5   & 43.5          & 88.3          & 92.7          & 94.4          \\
		PPLR+CAJ                 & -                          & \textbf{86.1} & \textbf{94.4} & \textbf{97.9} & \textbf{98.7} & \textbf{44.3} & \textbf{75.1} & \textbf{84.3} & \textbf{87.3} & \textbf{45.3} & {\ul 90.4}    & \textbf{93.9} & \textbf{95.2} \\ \hline
	\end{tabular}
	\label{sota}
\end{table*}

\section{Experiments}
\subsection{Datasets and Evaluation Protocols}
We evaluate the proposed method on two person re-ID datasets (Market1501 \cite{market1501}, MSMT17 \cite{PTGAN}) and one vehicle re-ID dataset (VeRi-776 \cite{veri}).
We adopt mean Average Precision (mAP) \cite{mAP} and Cumulative Matching Characteristic (CMC) \cite{CMC} to evaluate performance.

\subsection{Implement Details}
Our proposed CA-Jaccard distance can be applied in the clustering and re-ranking scenes of person re-ID. Therefore, to fully verify the effectiveness of our CA-Jaccard distance, we conduct experiments in both scenes.
CA-Jaccard distance can be applied with marginal modification. Specifically, only the Jaccard distance needs to be replaced with CA-Jaccard distance, while all other settings remain unchanged. In CA-Jaccard distance, we set the $k_1^{intra}$ and $k_1^{inter}$ to 5 and 20 in Eq.~(\ref{eqn11}). The $k_2^{intra}$ and $k_2^{inter}$ are set to 2 and 4 respectively in Eq.~(\ref{eqn12}). 

\subsection{Performance Improvement in Clustering Scene}
In Tab. \ref{sota}, We verify the effectiveness of our CA-Jaccard distance by applying it in state-of-the-art unsupervised person re-ID methods (\eg CAP \cite{CAP}, CC \cite{CC}, ICE\cite{ICE} and PPLR \cite{PPLR}). We can observe that when the CA-Jaccard distance is applied for clustering, the performance of these methods gains significant improvement. Especially when applying our CA-Jaccard distance to a more powerful method PPLR \cite{PPLR}, we achieve 86.1\%/94.4\% mAP/Rank-1 on Market1501, 44.3\%/75.1\% mAP/Rank-1 on MSMT17, and 45.3\%/90.4\% mAP/Rank-1 on VeRi-776, which surpasses all unsupervised person re-ID methods by a large margin. Moreover, we can find that CA-Jaccard distance can bring greater performance improvement on the MSMT17 and VeRi-776 datasets with larger camera variation compared to Market1501, demonstrating that CA-Jaccard distance effectively solves the problem of unreliable Jaccard distance caused by camera variation. The results show the effectiveness and generalization of our method.

\subsection{Peformance Improvement in Re-ranking Scene}
\begin{table}[ht]
	\renewcommand\tabcolsep{1.0pt}
	\renewcommand{\arraystretch}{1.0}
	\centering
	\caption{Comparison with the state-of-the-art re-ranking methods for person re-ID on Market1501, MSMT17 and VeRi-776. The best results are in \textbf{bold}.}
	\begin{tabular}{c|ccc|ccc|ccc}
		\hline
		\multirow{2}{*}{Methods} & \multicolumn{3}{c|}{Market1501}                & \multicolumn{3}{c|}{MSMT17}                    & \multicolumn{3}{c}{VeRi-776}                  \\
		& mAP           & R1            & R5            & mAP           & R1            & R5            & mAP           & R1            & R5            \\ \hline
		BoT~\cite{BoT}           & 85.9          & 94.5          & \textbf{98.2} & 50.7          & 74.0          & 85.6          & 76.2          & 95.5          & 98.1          \\
		+KR~\cite{jaccard}       & 94.2          & 95.4          & 97.9          & 66.9          & 79.4          & 86.6          & 78.7          & 95.8          & 97.2          \\
		+ECN~\cite{ECN}          & 94.4          & 95.9          & 97.8          & 69.0          & 80.5          & 86.3          & 79.5          & 96.8          & 97.3          \\ 
		+CAJ                     & \textbf{94.5} & \textbf{96.2} & 98.1          & \textbf{74.1} & \textbf{86.2} & \textbf{90.5} & \textbf{81.4} & \textbf{97.6} & \textbf{98.3} \\ \hline
		CC*~\cite{CC}            & 81.0          & 91.1          & 96.2          & 31.1          & 60.2          & 71.3          & 38.1          & 80.3          & 85.1          \\
		+KR~\cite{jaccard}       & 89.7          & 93.2          & 95.5          & 42.6          & 65.3          & 73.4          & 39.0          & 80.0          & 81.6          \\
		+ECN~\cite{ECN}          & 90.0          & 93.4          & 95.0          & 43.9          & 65.3          & 71.8          & 40.0          & 81.1          & 82.4          \\ 
		+CAJ                     & \textbf{90.2} & \textbf{93.7} & \textbf{95.9} & \textbf{45.3} & \textbf{68.9} & \textbf{75.3} & \textbf{42.5} & \textbf{88.6} & \textbf{91.1} \\ \hline
	\end{tabular}
	\label{sota_rerank}
\end{table}
We apply CA-Jaccard to re-ranking the retrieval results of pre-trained models of supervised and unsupervised commonly used baselines (BoT \cite{BoT} and CC \cite{CC}). For a fair comparison, we also apply the state-of-the-art re-ranking methods \ie KR~\cite{jaccard}, and ECN~\cite{ECN}. The experiment results are reported in Tab. \ref{sota_rerank}.
We can observe that our method improves the performance of BoT and CC by a large margin. Meanwhile, our method consistently brings greater performance improvement than the state-of-the-art re-ranking methods. These results demonstrate the effectiveness and superiority of our method.
\subsection{Ablation Studies}

\begin{figure*}[ht]
	\centering
	\includegraphics[width=0.84\linewidth]{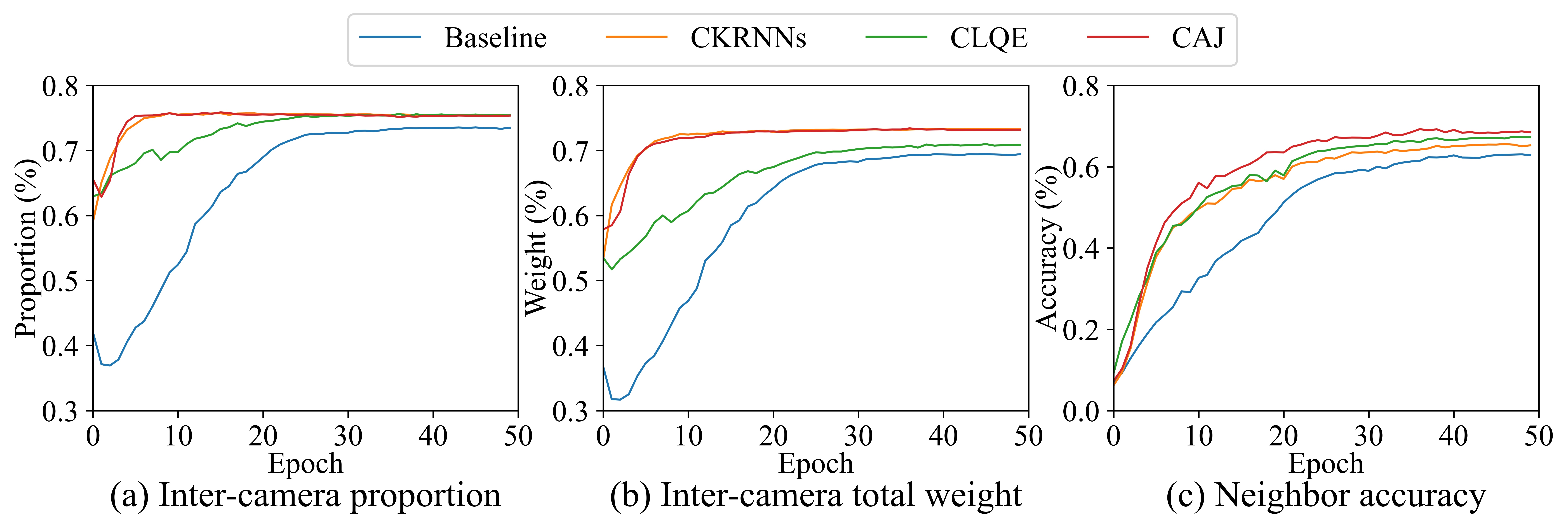}
	\caption{(a) average inter-camera proportion, (b) average inter-camera total weight and (c) average neighbor accuracy of all training samples' weighted expanded neighbors vectors over different epochs from baseline, CKRNNs, CLQE and CAJ in clustering scene.}
	\label{pwm}
\end{figure*}
In this section, we conduct extensive experiments on Maret1501 and MSMT17 in clustering and re-ranking scenes to validate the effectiveness of each component in our method. We select CC with instance memory updating mechanism and average pooling as the baseline for clustering scene and BoT as the baseline for re-ranking scene. We present the results of the baselines and three variants of our CA-Jaccard distance in Tab. \ref{ablation}. 
Then we analyze each component of our method respectively.
\begin{table}[t]
	\caption{Ablation study on individual components in the clustering and re-ranking scenes. ``CAJ" represents CA-Jaccard distance.}
	\centering
	\renewcommand\tabcolsep{1.9pt}
	\renewcommand{\arraystretch}{0.9}
	\begin{tabular}{c|ccc|ccc}
		\hline
		\multirow{2}{*}{Method} & \multicolumn{3}{c}{Market1501} & \multicolumn{3}{|c}{MSMT17} \\
		& mAP     & Rank-1    & Rank-5   & mAP    & Rank-1  & Rank-5  \\ \hline
		\multicolumn{7}{c}{Clustering scene}                                                  \\ \hline
		CC*~\cite{CC}           & 81.0    & 91.1      & 96.2     & 31.1   & 60.2    & 71.3    \\
		+CKRNNs                 & 82.4    & 92.9      & 97.1     & 35.1   & 65.3    & 75.8    \\
		+CLQE                   & 83.5    & 92.6      & 97.0     & 40.4   & 70.1    & 81.2    \\
		+CAJ                    & \textbf{84.8}    & \textbf{93.6}      & \textbf{97.6}     & \textbf{42.8}   & \textbf{72.3}    & \textbf{82.2}    \\ \hline
		\multicolumn{7}{c}{Re-ranking scene}                                                  \\ \hline
		BoT~\cite{BoT}          & 85.9    & 94.5      & \textbf{98.2}     & 50.7   & 74.0    & 85.6    \\
		BoT+KR\cite{jaccard}   & 94.2     & 95.4      & 97.9     & 66.9   & 79.4    & 86.6    \\
		+CKRNNs              & 94.4    & 95.7      & 97.9     & 73.1   & 84.0    & 90.3    \\
		+CLQE                & 94.3    & 95.7      & 98.0     & 72.0   & 85.5    & 90.2    \\
		+CAJ                 & \textbf{94.5}    & \textbf{96.2}      & 98.1     & \textbf{74.1}   & \textbf{86.2}    & \textbf{90.5}     \\ \hline
	\end{tabular}
	\label{ablation}
	
\end{table}

\textbf{Effect of CKRNNs.}
To verify the effectiveness of CKRNNs, we replace the robust KRNNs in Jaccard distance with CKRNNs. The results in Tab. \ref{ablation} demonstrate a significant performance improvement compared to the baselines. Specifically, in the clustering scene, applying CKRNNs brings 1.4\% mAP and 1.8 \% Rank-1 improvement on Market1501, and 4.0\% mAP and 5.1\% Rank-1 improvement on MSMT17. 
In the re-ranking scene, CKRNNs consistently improve the performance of BoT+KR. 
Especially on the challenging MSMT17 datasets, it brings 6.2\%/4.6\% mAP/Rank-1 improvement.
These results validate the effectiveness of the CKRNNs.

\textbf{Effect of CLQE.}
To validate the necessity of CLQE, we incorporate CLQE into the Jaccard distance. The experimental results, presented in Tab. \ref{ablation}, show that CLQE provides a significant performance improvement in both scenarios. CLQE improves the mAP and Rank-1 by 2.5\% and 1.5\% on Market1501, and 9.3\% and 9.9\% on MSMT17 respectively in the clustering scenario.
Meanwhile, in the re-ranking scene, mAP/Rank-1 are improved
when CLQE is applied.
These results underscore the importance of CLQE in effectively mining reliable samples and increasing the weights of reliable samples.

\textbf{Neighbors analysis.}
To further investigate the effectiveness of CKRNNs and CLQE, we plot three line charts in Fig. \ref{pwm}, which represent the average inter-camera proportion, average inter-camera total weight, and average neighbor accuracy of all training samples' weighted expanded neighbors vectors over different epochs from baseline, CKRNNs, CLQE and CA-Jaccard distance in clustering scene. As shown in Fig. \ref{pwm} (a) and (b), we can observe that CKRNNs and CLQE improve the average proportion and total weight of inter-camera samples in the weighted expanded neighbors vectors. However, the combination of CKRNNs and CLQE results in a subtle difference in proportion and weight compared to using CKRNNs alone. This can be attributed to the fact that most of the inter-camera samples brought by CLQE are already included in CKRNNs. Therefore, when CKRNNs and CLQE are used together, CKRNNs mainly focus on improving the proportion and total weight of inter-camera samples in relevant neighbors, while CLQE focuses more on improving the weights of reliable samples. Moreover, Fig. \ref{pwm} (c) demonstrates that the simultaneous use of CKRNNs and CLQE leads to better average neighbor accuracy compared to using either one alone. This suggests that CA-Jaccard distance maximizes the reliability of relevant neighbors and distance, resulting in performance improvement.
\begin{figure*}
	\centering
	\includegraphics[width=0.83\linewidth]{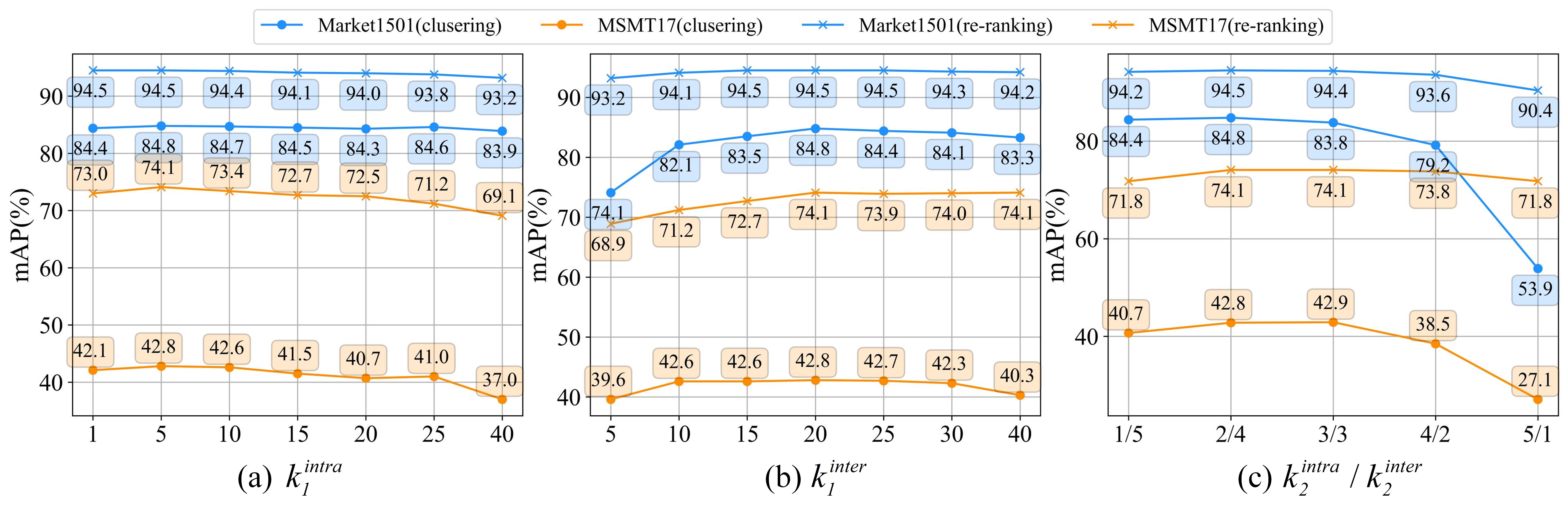}
	\caption{Parameter analysis of $k_1^{intra}$, $k_1^{inter}$ and $k_2^{intra}/k_2^{inter}$ on Market1501 and MSMT17.}

	\label{parameter}
\end{figure*}

\subsection{Parameter Analysis}
In CA-Jaccard distance, four parameters are introduced, including $k_1^{intra}$, $k_1^{inter}$ for CKRNNs and $k_2^{intra}$, $k_2^{inter}$ for CLQE. We conduct experiments to analyze the impact of each parameter on Market1501 and MSMT17 datasets in both clustering and re-ranking scene. CC and BoT are the baselines for clustering and re-ranking scene. The mAP results are presented in Fig. \ref{parameter}.

\textbf{Impact of $k_1^{intra}$ and $k_1^{inter}$ in CKRNNs.}
In Fig. \ref{parameter} (a) and (b), we investigate the impact of $k_1^{intra}$ and $k_1^{inter}$. We observe that the performance remains stable when $k_1^{intra}$ is within the range from 1 to 20 and $k_1^{inter}$ is within the range from 15 to 30. This is because CLQE decreases the impact of $k_1^{intra}$ and $k_1^{inter}$ in CKRNNs by emphasizing reliable samples in weighted expanded neighbors vectors. However, when $k_1^{inter}$ is set to 5, there is a significant decrease in performance. Conversely, setting $k_1^{intra}$ to 1, meaning that the intra-camera neighbors of samples only include themselves, still achieves high performance. This finding validates that inter-camera samples have more information and reliability than intra-camera samples.  Moreover, we find that setting $k_1^{intra}$ or $k_1^{inter}$ with too large values will bring too many noise samples and hinder the performance. Therefore, considering the performance on two datasets, we set $k_1^{intra}$ to 5 and $k_1^{inter}$ to 20.

\textbf{Impact of $k_2^{intra}$ and $k_2^{inter}$ in CLQE.}
Due to intra-camera and inter-camera k-nearest neighbors being noisy, we follow \cite{jaccard} and limit the sum of $k_2^{intra}$ and $k_2^{inter}$ to 6. In Fig. \ref{parameter} (c), we vary $k_2^{intra}$/$k_2^{inter}$ from 1/5 to 5/1. A smaller $k_2^{intra}$ and a larger $k_2^{inter}$ lead to the disregard of intra-camera information, thereby limiting the performance. Meanwhile, too large $k_2^{intra}$ and too small $k_2^{inter}$ weaken the mining ability of CLQE for reliable samples and lead to a decrease in performance. These experimental results lead us to set $k_2^{intra}$ = 2 and $k_2^{inter}$ = 4.

\subsection{Visualizations}
To better understand the effect of our CA-Jaccard distance, we conduct visualizations to qualitatively analyze the impact of CA-Jaccard distance.

\begin{figure}
	\centering
	\includegraphics[width=0.42\textwidth]{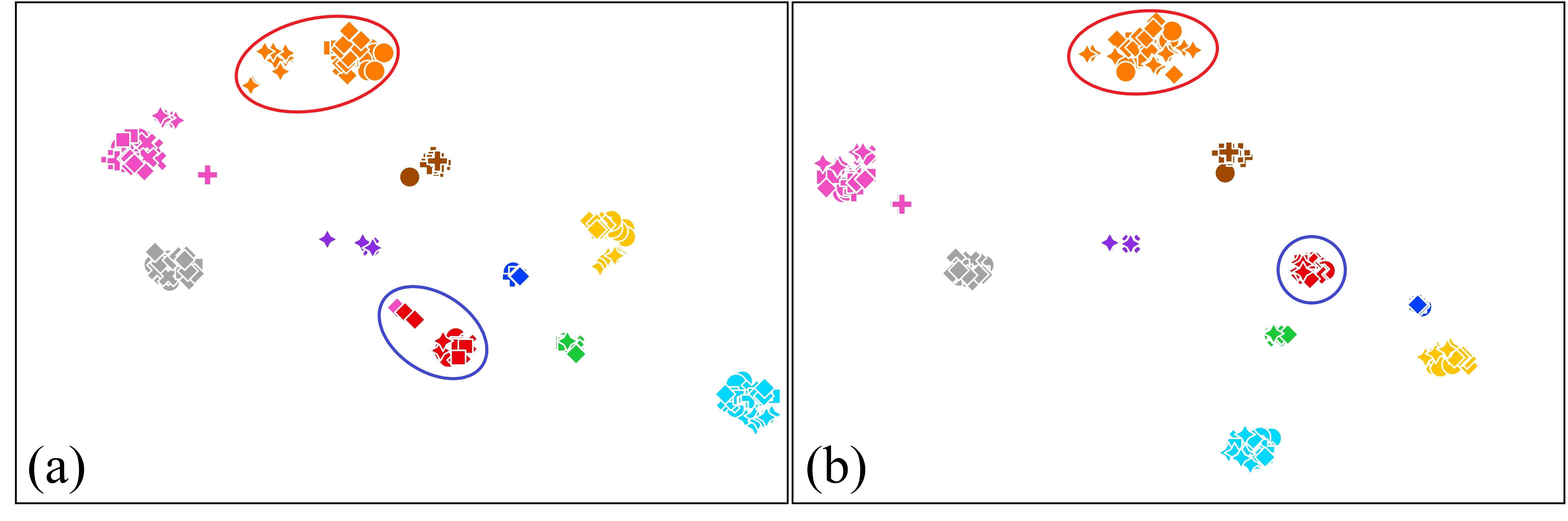}
	\caption{The t-SNE visualization of 10 persons' features extracted by the models of (a) CC and (b) CC+CAJ. Different colors and shapes indicate different identities and camera labels.}
	\label{tsne}
\end{figure}

\textbf{Clustering scene.} We make t-SNE visualization \cite{tsne} on Market1501. 
As illustrated in Fig. \ref{tsne}, our method compacts the samples of same person from different cameras (\eg red circle and blue circle in Fig. \ref{tsne}), indicating that our CA-Jaccard distance helps generate more accurate pseudo labels that guide the model learning camera-invariant features.

\begin{figure}
	\centering
	\includegraphics[width=0.4\textwidth]{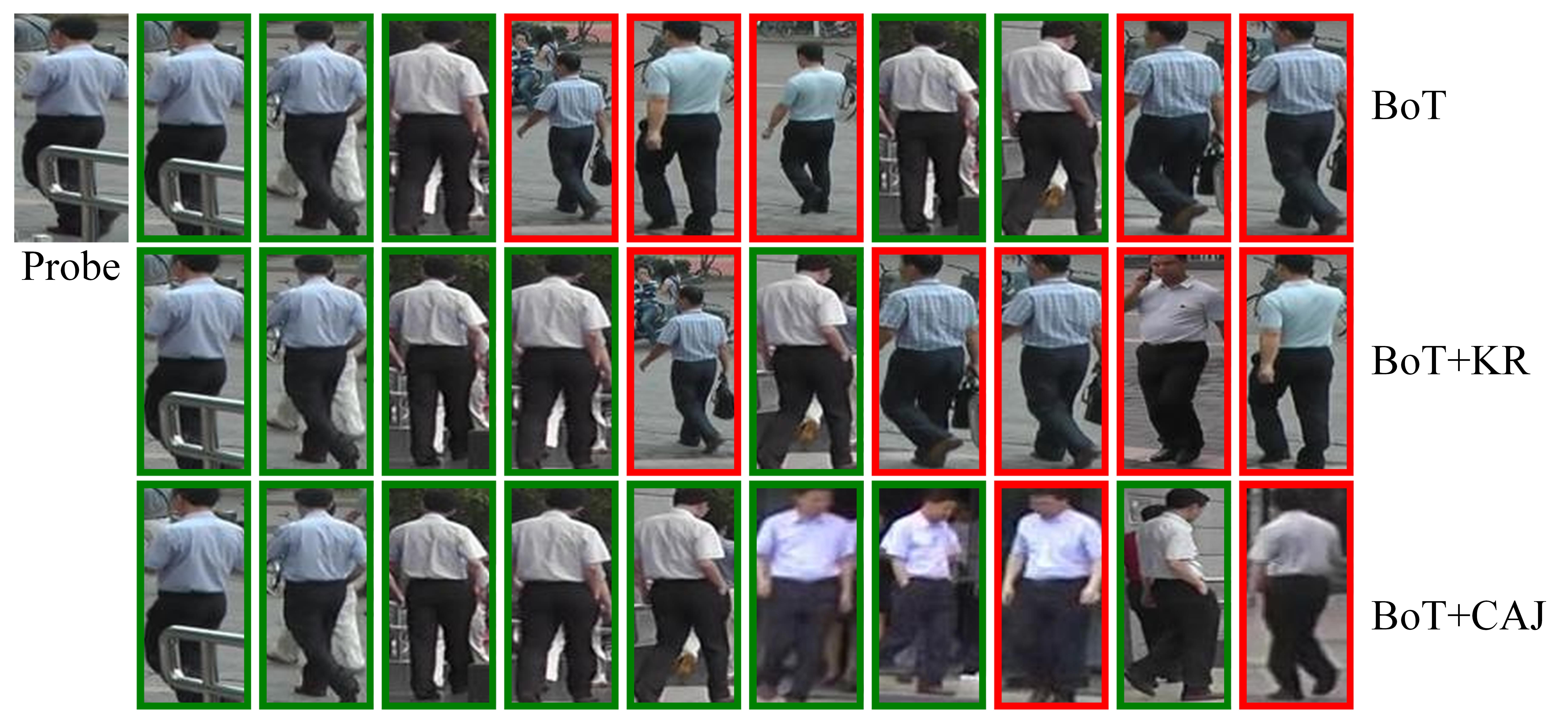}
	\caption{
		Ranking results of a probe produced by BoT, BoT+KR and BoT+CAJ respectively.
	}
	\label{retrieval_zhenwen}
\end{figure}

\textbf{Re-ranking scene.} Ranking results of BoT, BoT+KR, BoT+CAJ on Market1501 are represented in Fig. \ref{retrieval_zhenwen}. Compared to KR re-ranking that uses Jaccard distance for re-ranking, CA-Jaccard distance achieves better ranking results, which indicates the superiority of our method.



\subsection{Computational Complexity Analysis}
We replace the robust KRNNs and LQE in the Jaccard distance with CKRNNs and CLQE, while keeping other parts consistent. The computation of CKRNNs includes sorting and applying k-reciprocal nearest constraint. Thus the computational complexity of CKRNNs is $O(N^{2}logN)$, which is comparable to that of KRNNs. However, robust KRNNs still have a recall operation which is a time-consuming set operation. The computational complexity of CLQE stays the same as LQE. In summary, CA-Jaccard distance has lower computational complexity and more reliable distance than the Jaccard distance.
\section{Conclusion}

In this paper, we propose a novel CA-Jaccard distance for person re-ID that overcomes camera variation and enhances the reliability of Jaccard distance through the use of CKRNNs and CLQE. CKRNNs improve reliability by incorporating informative inter-camera positive samples while excluding intra-camera negative samples in neighbors. CLQE mines reliable samples in CKRNNs and assigns higher weights to them to further enhance the reliability. Extensive ablation studies and experiment results validate the effectiveness and robustness of our method. The low computational complexity and effectiveness of our CA-Jaccard distance make it a general distance metric for person re-ID.
\newpage
{
    \small
    \bibliographystyle{ieeenat_fullname}
    \bibliography{main}
}

\clearpage
\setcounter{page}{1}
\maketitlesupplementary

\section{The Computation Details of CA-Jaccard Distance}
Alg. \ref{alg1} delineates the whole computation steps for CA-Jaccard distance. 
First, extract features of all samples by model $f_\theta$ and calculate the original distance matrix $D$.
Then, find CKRNNs by applying k-reciprocal constraint in intra-camera and inter-camera ranking lists.
Next, turn these CKRNNs of samples into weighted CKRNNs vectors.
Subsequently, use CLQE to obtain the weighted neighbors vectors.
Finally, CA-Jaccard distance matrix $D^{CAJ}$ is computed by the overlap between weighted expanded neighbors vectors of samples.
 
\begin{algorithm}[H]
	\renewcommand{\algorithmicrequire}{\textbf{Input:}}
	\renewcommand{\algorithmicensure}{\textbf{Output:}}
	\footnotesize
	\caption{The computation procedures of CA-Jaccard distance}
	\label{alg1}
	\begin{algorithmic}[1]
			\REQUIRE Dataset $X = \{ {x_i}\} _{i = 1}^{{N}}$ with camera labels $\{ {c_i}\} _{i = 1}^{{N}}$, model $f_\theta$;
			\ENSURE Jaccard distance matrix $D^{CAJ}$;
			\STATE Extract features from $X$ by the model $f_\theta$;
			\STATE {\color{gray}// CA-Jaccard distance computation steps}
			\STATE {\color{gray}// Step 1: Original distance computation}
			\STATE Calculate original distance matrix $D$;
			\STATE {\color{gray}// Step 2: CKRNNs}
			\STATE Find CKRNNs with Eq.~\ref{eqn11};
			\STATE {\color{gray}// Step 3: Vectorization of neighbors}
			\STATE Encode CKRNNs into weighted CKRNNs vectors with Eq.~\ref{eqn4};
			\STATE {\color{gray}// Step 4: CLQE}
			\STATE Use CLQE (Eq.~\ref{eqn12}) to obtain weighted expanded neighbors vectors
			\STATE {\color{gray}//  Step 5: Overlap computation}
			\STATE Compute the CA-Jaccard distance matrix $D^{CAJ}$ with Eq.~\ref{eqn8};
			
		\end{algorithmic}
\end{algorithm}

\section{Additional Visualizations}
Some additional visualizations are presented to further verify the effectiveness of CA-Jaccard distance.

\begin{figure}
	\centering
	\includegraphics[width=0.4\textwidth]{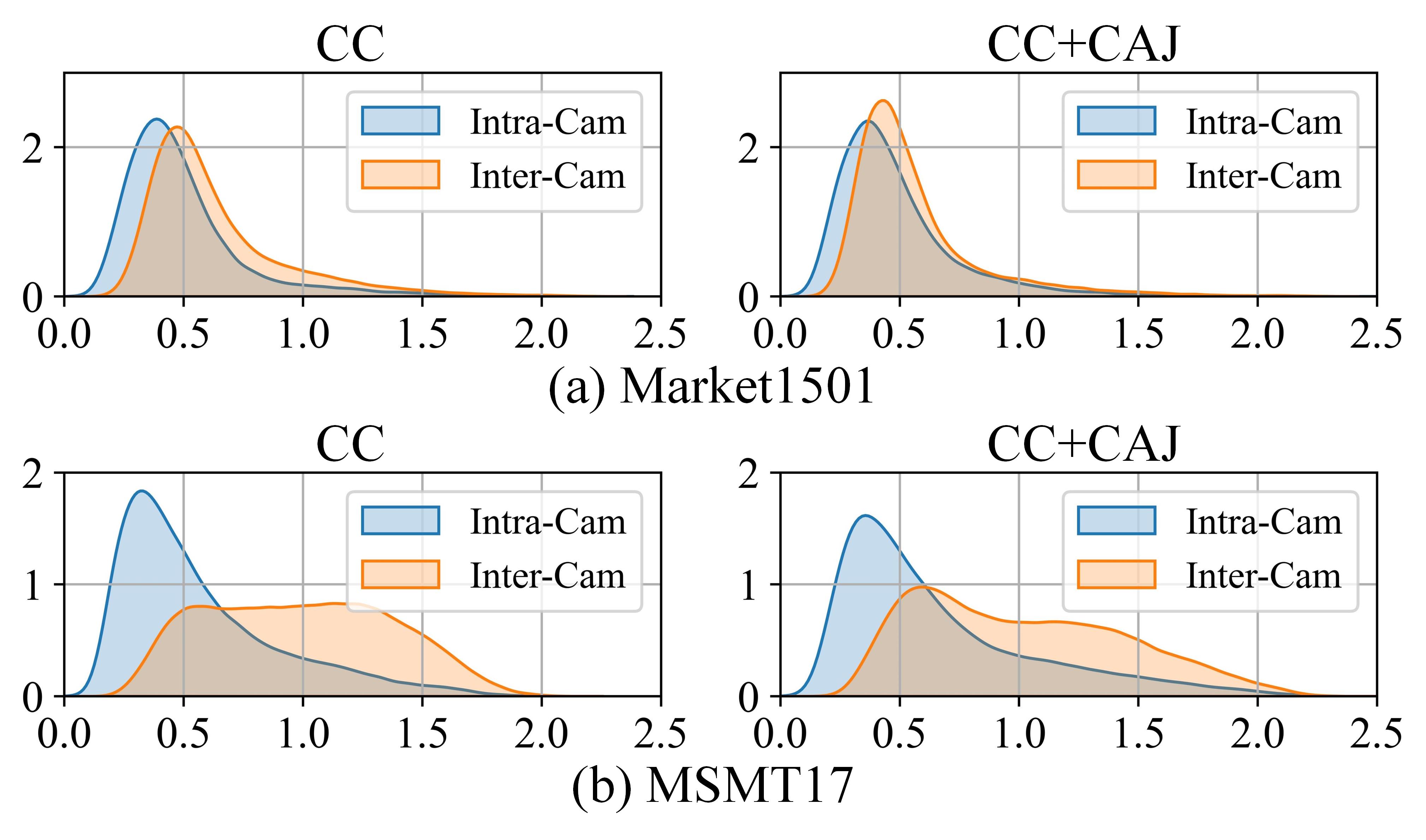}
	\caption{Distance distributions of intra-camera and inter-camera positive pairs on (a) Market1501 and (b) MSMT17.}
	\label{distribution}
\end{figure}

\textbf{Clustering scene.} 
We visualize the distance distribution of intra-camera and inter-camera positive pairs for two datasets in Fig. \ref{distribution}. 
As shown in  Fig. \ref{distribution}, compared to baseline, our CA-Jaccard distance significantly reduces the difference between distribution of intra-camera and inter-camera positive pairs for both datasets. 
These observations further verify the effectiveness and reliability of our CA-Jaccard distance.

\begin{figure}
	\centering
	\includegraphics[width=0.48\textwidth]{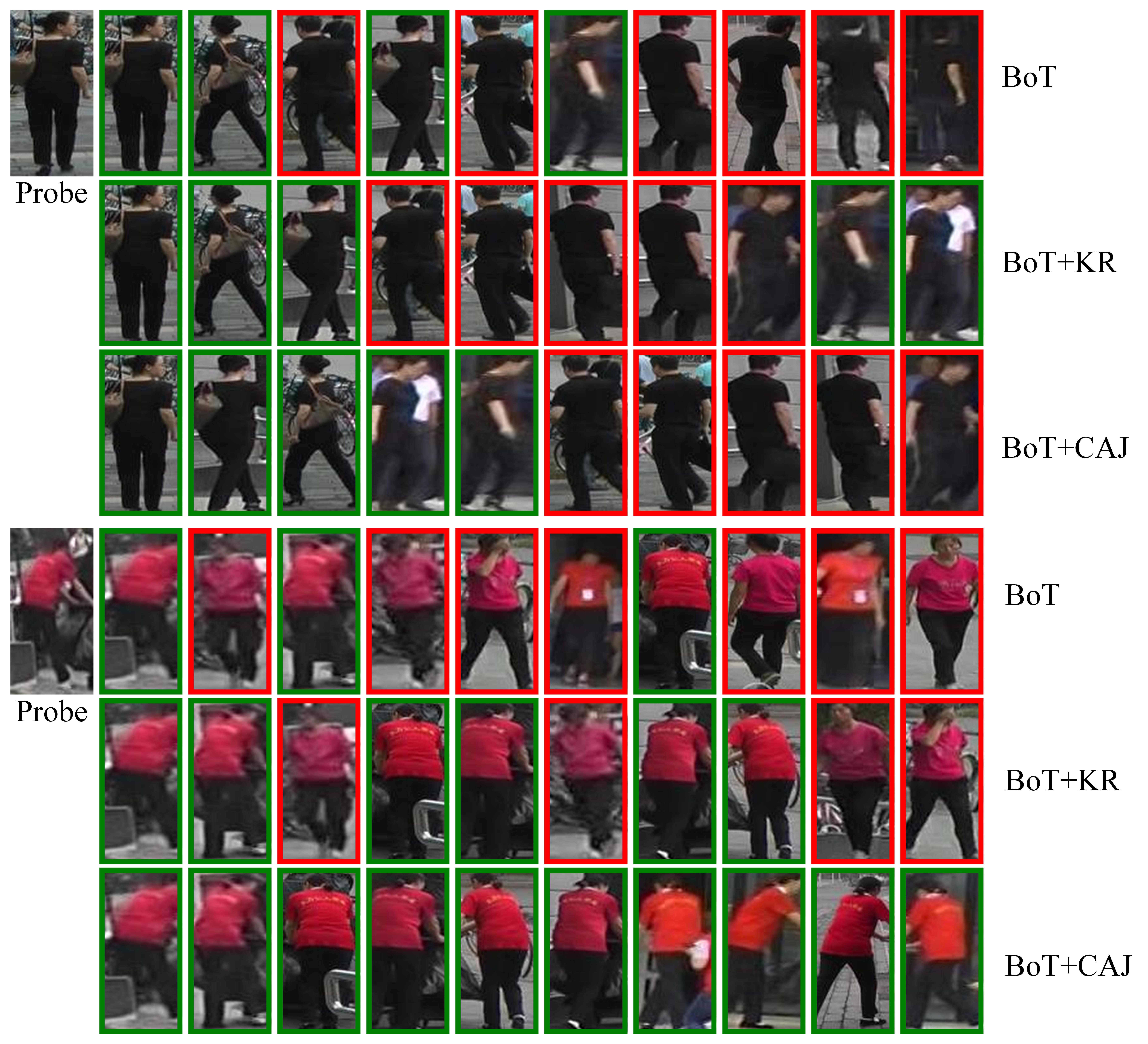}
	\caption{More retrieval results on the Market1501 dataset. For each probe, the first, second, and third rows correspond to the ranking results produced by BoT, BoT+KR, and BoT+CAJ respectively. The person surrounded by a green box represents the same person as the probe, while the person surrounded by a red border represents a different person from the probe image.}
	\label{retrieval_supply}
\end{figure}

\textbf{Re-ranking scene.}
More retrieval results are visualized in Fig.~\ref{retrieval_supply}.
CA-Jaccard distance effectively ranks more positive samples into the top of ranking list which are absent in the ranking lists of BoT and BoT+KR. These findings indicate that CA-Jaccard distance metric can produce more accurate distances.

\end{document}